\documentclass{article}
\usepackage{spconf,amsmath,graphicx}

\usepackage {colortbl,array,xcolor}
\newcommand{\bhline}[1]{\noalign{\hrule height #1}}  

\newcolumntype{C}{>{\centering\arraybackslash}X}
\usepackage{booktabs}
\usepackage{multirow}
\usepackage{bm}
\usepackage{wrapfig}
\usepackage{url}
\usepackage{tabularx}

\usepackage{enumitem}
\setlist{nosep, leftmargin=14pt}

\usepackage{mwe} 

\title{Hyperparameter-Free Boundary-Interior t-vMF Dice Loss\\ for Medical Image Segmentation}
\name{Hinako Mitsuoka$^{\dagger}$ \qquad Kazuhiro Hotta$^{\ddagger}$
}
\address{Meijo University, 1-501 Shiogamaguchi, Tempaku-ku, Nagoya, 468-8502, Japan \\
$^{\dagger}$263441509@ccmailg.meijo-u.ac.jp \qquad $^{\ddagger}$kazuhotta@meijo-u.ac.jp
}
\begin{document}
%
\maketitle
%
%
%
\begin{abstract}\
Class imbalance and errors near anatomical boundaries remain challenges in medical image segmentation, motivating loss functions that account for these difficulties. Adaptive t-vMF Dice Loss adjusts the similarity function during training but requires hyperparameter tuning. We propose Self-Tuned Boundary-Aware t-vMF Dice Loss (STB-tvMF Dice), which adapts the function separately for boundary and interior regions without tuning shape-adaptation parameters. Each region's shape parameter of t-vMF function is determined from its mean prediction-target cosine similarity in the preceding epoch and remains fixed throughout the current epoch. This rule maximizes the slope of the similarity function at the observed mean similarity, providing a principled basis for adaptation without validation-based parameter selection. Experiments on three datasets and architectures demonstrated its effectiveness. On Synapse with U-Net, STB-tvMF Dice achieved a mean DSC of 70.75\%, compared with 63.03\% for Adaptive t-vMF Dice Loss. Our code will be available soon.
\end{abstract}
\section{Introduction}
\label{sec:intro}

Medical image segmentation is challenged by substantial class imbalance and ambiguous image information near anatomical boundaries. Loss functions play an important role in determining how prediction errors are evaluated during training. 
Cross-entropy and Dice losses\cite{dice} provide complementary objectives based on pixel-wise classification and region overlap, respectively. 
Various extensions have been developed to address their limitations. 
Tversky loss adjusts the relative contributions of false positives and false negatives, while Focal Tversky loss emphasizes difficult predictions\cite{tversky, focaltversky}. 
Other approaches combine cross-entropy and Dice through additive or multiplicative formulations to incorporate both objectives\cite{multi}. 
Beyond selecting which errors to emphasize, the mapping from prediction-target agreement to a loss value determines how improvements in agreement influence optimization. Adaptive t-vMF Dice Loss addresses this aspect by adapting a similarity function using class-wise validation performance\cite{adaptivetvmf}. 
However, its adaptation requires validation data and a manually selected scaling coefficient. Moreover, prediction agreement can differ between boundary and interior regions within the same class, motivating adaptation that distinguishes these regions.
This raises the question of whether the similarity function can be determined directly from the learning state of each region without manual parameter search.

\begin{figure}[t!]
\begin{center}
\includegraphics[scale=0.275]{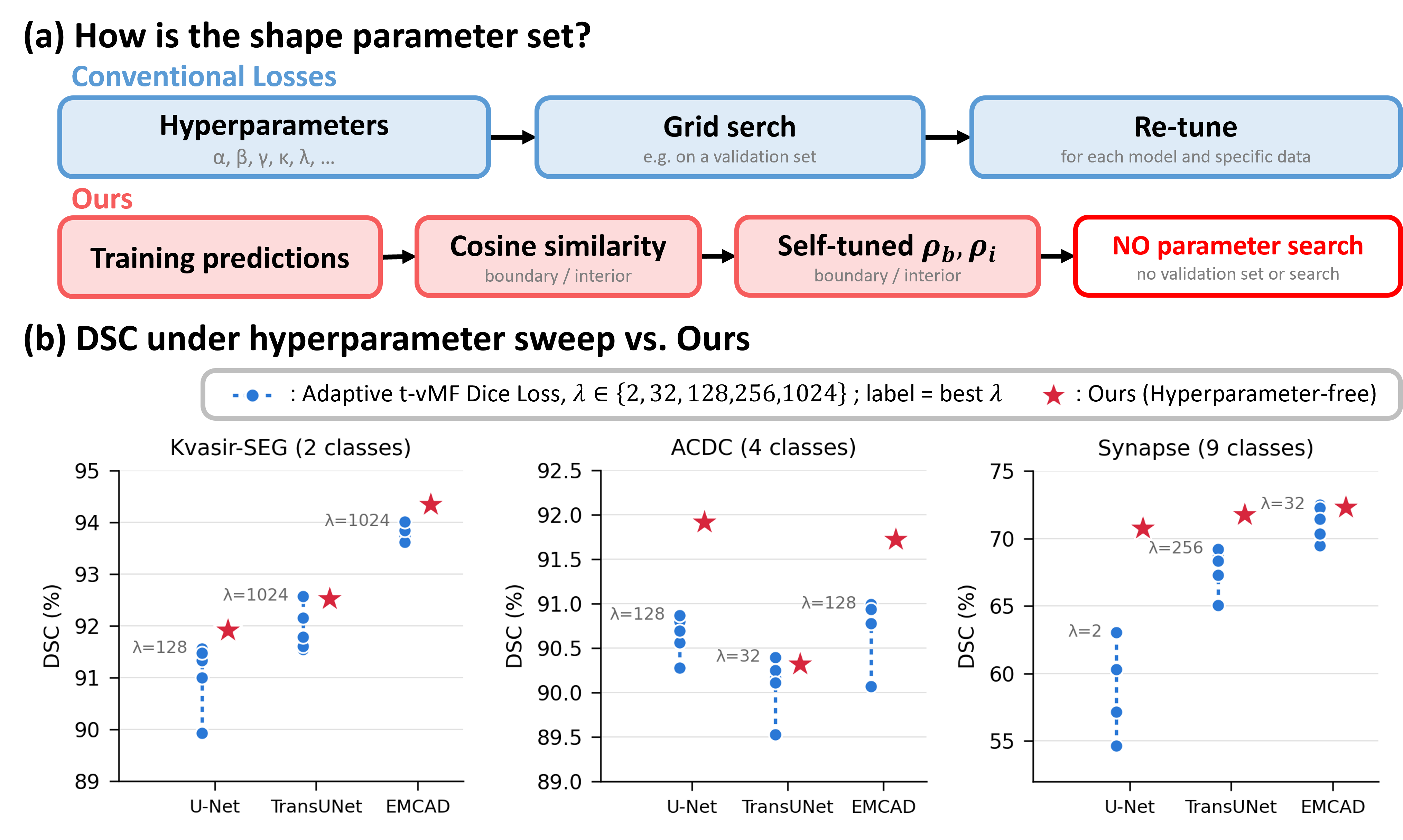}
\end{center}
\caption{
Self-tuning mechanism and performance of STB-tvMF Dice Loss.
(a) Boundary and interior shape parameters are updated using mean cosine similarities from the previous training epoch, without hyperparameter search.
(b) DSC across three datasets and three architectures.
Blue dots show Adaptive t-vMF Dice Loss with $\lambda \in \{2, 32, 128, 256, 1024\}$; dotted lines connect these settings, not uncertainty intervals.
Labels indicate the best $\lambda$ selected retrospectively by test DSC.
Red stars denote our method.
}
\label{fig:tr}
\end{figure}

We propose Self-Tuned Boundary-Aware t-vMF Dice Loss (STB-tvMF Dice).
The proposed loss partitions pixels into boundary and interior regions using ground-truth labels and evaluates prediction-target cosine similarity separately in each region. 
For each region, a bounded parameter $\rho$ controls the shape of the similarity function. We derive a self-tuning rule by maximizing the slope of this function at the observed mean cosine similarity.
The resulting rule sets $\rho$ to the mean cosine similarity of the corresponding region. 
Statistics are accumulated during training, and the mean from the preceding epoch determines the shape parameter for the current epoch. 
This procedure adapts the shape parameters using training statistics alone, without validation data or hyperparameter search.

Our contributions are twofold.
First, we introduce a boundary-interior t-vMF Dice formulation with a self-tuning rule grounded in the local sensitivity of the similarity function, eliminating manual search for loss-specific adaptation parameters.
Second, we evaluated the proposed loss through experiments on Kvasir-SEG, ACDC, and Synapse\cite{kvasir, acdc, smo} using U-Net, TransUNet, and EMCAD\cite{unet, transunet, emcad}.
Compared with the best tested scaling-coefficient settings of Adaptive t-vMF Dice Loss, our method improved the mean Dice similarity coefficient (DSC) on Synapse by 7.72 and 2.56\% with U-Net and TransUNet, respectively.
With EMCAD, our method outperformed all compared losses on Kvasir-SEG and ACDC, and achieved a comparable DSC on Synapse for the best Adaptive t-vMF setting.
Our method achieved these results without manual search for shape-adaptation parameters.

\section{Methodology}
\label{sec:method}

We propose Self-Tuned Boundary-Aware t-vMF Dice Loss (STB-tvMF Dice). Unlike Adaptive t-vMF Dice Loss\cite{adaptivetvmf}, which uses class-wise validation performance and a manually selected scaling coefficient, our method determines separate boundary and interior shape parameters directly from training similarities.

\subsection{Boundary-Interior Similarity}
\label{sec:regional_similarity}

Let $p_{n,k}(u)$ denote the predicted probability of class $k$ at pixel $u$ in image $n$, and let $y_{n,k}(u)$ be the corresponding one-hot ground-truth label.
We identify boundary pixels where the ground-truth label differs from its local average computed by $3\times3$ average pooling with stride one and zero padding; the remaining pixels form the interior region.
The resulting binary masks are denoted by $m_{n,b}$ and $m_{n,i}$ and are shared across classes.
The neighborhood size is fixed across all datasets, and any ignored pixels are excluded from both masks.

For each region $r\in\{b, i\}$, we compute cosine similarity between the masked prediction and target:
\begin{equation}
x_{n,k,r}
=
\left\langle
\mathcal{N}(p_{n,k}\odot m_{n,r}),
\mathcal{N}(y_{n,k}\odot m_{n,r})
\right\rangle,
\label{eq:regional_cosine}
\end{equation}
where $\odot$ denotes element-wise multiplication and $\mathcal{N}(v)=v/\max(\|v\|_2,\epsilon_{\mathrm{norm}})$.
The norm and inner product are computed over spatial positions.
Boundary and interior agreement are thus evaluated separately, with one shape parameter per region shared across classes.

\subsection{Bounded Parameterization and Self-Tuning}
\label{sec:self_tuning}

The t-vMF similarity\cite{tvmf} function transforms cosine similarity $x$ using a nonnegative concentration parameter $\kappa$ that controls its shape as
\begin{equation}
\phi_{\mathrm{t}}(x;\kappa)
=
\frac{1+x}{1+\kappa(1-x)}-1.
\label{eq:tvmf_original}
\end{equation}
where $x$ is the regional cosine similarity defined in Eq.~\eqref{eq:regional_cosine}. When $\kappa=0$, the function reduces to cosine similarity. We introduce the bounded shape parameter $\rho=\kappa/(1+\kappa)\in[0,1)$. Substituting $\kappa=\rho/(1-\rho)$ into Eq.~\eqref{eq:tvmf_original} yields
\begin{equation}
\phi(x;\rho)
=
\frac{x-\rho}{1-\rho x}.
\label{eq:tvmf_rho}
\end{equation}
This equivalent parameterization avoids explicitly computing an unbounded $\kappa$ and ensures a positive denominator for $x\in[0,1]$ and $\rho<1$.

Let $c$ denote the mean prediction-target cosine similarity within a region, averaged over valid training image-class pairs as defined in Eq.~\eqref{eq:epoch_mean}. 
We determine $\rho$ by maximizing the local sensitivity of the similarity function at $x=c$, measured by its slope

\begin{equation}
S(\rho;c)
=
\left.
\frac{\partial\phi(x;\rho)}{\partial x}
\right|_{x=c}
=
\frac{1-\rho^2}{(1-\rho c)^2}.
\label{eq:sensitivity}
\end{equation}
Differentiating with respect to $\rho$ gives
\begin{equation}
\frac{\partial S(\rho;c)}{\partial\rho}
=
\frac{2(c-\rho)}{(1-\rho c)^3}.
\label{eq:sensitivity_derivative}
\end{equation}
For $0\leq c<1$, this derivative is positive for $\rho<c$ and negative for $\rho>c$.
Therefore, the maximizing parameter is
\begin{equation}
\rho^\star=c.
\label{eq:optimal_rho}
\end{equation}
This criterion maximizes the slope of the similarity function at the regional mean, rather than the gradient magnitude of the complete loss.

\begin{table*}[t]
\centering
\caption{DSC (\%) on three datasets using three architectures.
The best result in each column is shown in bold.}
\label{tab:main_results}
\fontsize{9}{11}\selectfont
\setlength{\tabcolsep}{3pt}
\renewcommand{\arraystretch}{1.05}
\begin{tabular*}{\textwidth}{@{\extracolsep{\fill}}l*{9}{c}@{}}
\bhline{1.5pt}
& \multicolumn{3}{c}{U-Net \cite{unet}}
& \multicolumn{3}{c}{TransUNet \cite{transunet}}
& \multicolumn{3}{c}{EMCAD \cite{emcad}} \\
\cmidrule(lr){2-4}
\cmidrule(lr){5-7}
\cmidrule(lr){8-10}
Loss
& Kvasir & ACDC & Synapse
& Kvasir & ACDC & Synapse
& Kvasir & ACDC & Synapse \\
\hline
Cross-entropy
& 89.68 & 90.07 & 48.37
& 92.55 & 86.89 & 45.89
& 93.46 & 88.04 & 62.50 \\
Dice \cite{dice}
& 88.98 & 89.67 & 65.02
& 92.65 & 89.58 & 68.19
& 93.50 & 90.04 & 71.10 \\
Tversky \cite{tversky}
& 89.28 & 89.86 & 57.52
& 92.40 & 90.31 & 52.41
& 93.65 & 90.04 & 67.97 \\
Focal Tversky \cite{focaltversky}
& 88.33 & 89.18 & 56.93
& 92.32 & 89.76 & 59.21
& 93.10 & 89.86 & 69.08 \\
CE + Dice
& 90.23 & 90.82 & 65.20
& 92.17 & 89.42 & 66.21
& 92.40 & 89.98 & 70.57 \\
CE $\times$ Dice
& 86.41 & 87.83 & 48.51
& 91.97 & 85.29 & 47.04
& 92.44 & 85.91 & 62.26 \\
TopK Dice \cite{topkdice}
& 90.84 & 90.59 & 65.28
& 92.11 & 89.54 & 68.12
& 93.63 & 89.62 & 69.45 \\
ABeDice \cite{abe}
& 89.09 & 89.25 & 61.97
& \textbf{92.72} & 87.83 & 65.10
& 93.78 & 88.90 & 67.29 \\
\hline
Adaptive t-vMF Dice \cite{adaptivetvmf} ($\lambda=2$) 
& 89.92 & 90.56 & 63.03
& 92.15 & 89.53 & 65.04
& 93.89 & 90.07 & 69.48 \\
Adaptive t-vMF Dice ($\lambda=32$)
& 91.33 & 90.80 & 60.31
& 91.55 & \textbf{90.40} & 67.30
& 93.74 & 90.96 & \textbf{72.48} \\
Adaptive t-vMF Dice ($\lambda=128$)
& 91.56 & 90.86 & 57.17
& 91.61 & 90.16 & 68.65
& 93.62 & 90.99 & 72.24 \\
Adaptive t-vMF Dice ($\lambda=256$)
& 91.48 & 90.28 & 60.32
& 91.78 & 90.25 & 69.19
& 93.85 & 90.78 & 71.42 \\
Adaptive t-vMF Dice ($\lambda=1024$)
& 91.00 & 90.69 & 54.64
& 92.57 & 90.25 & 68.35
& 94.01 & 90.94 & 70.35 \\
\hline
STB-tvMF Dice (Ours)
& \textbf{91.91} & \textbf{91.92} & \textbf{70.75}
& 92.52 & 90.32 & \textbf{71.75}
& \textbf{94.33} & \textbf{91.72} & 72.29 \\
\bhline{1.5pt}
\end{tabular*}
\end{table*}

\subsection{Epoch-Wise Adaptation and Training Objective}
\label{sec:epoch_adaptation}

During epoch $e$, we average regional cosine similarities as
\begin{equation}
c_r^{(e)}
=
\frac{1}{|\mathcal{V}_r^{(e)}|}
\sum_{(n,k)\in\mathcal{V}_r^{(e)}} x_{n,k,r},
\label{eq:epoch_mean}
\end{equation}
where $\mathcal{V}_r^{(e)}$ includes training image-class combinations with at least one ground-truth pixel of class $k$ in region $r$. 
Each combination is weighted equally, regardless of pixel count.

The next epoch uses
\begin{equation}
\rho_r^{(e+1)}
=
\min\!\left(\max(c_r^{(e)},0),1-\epsilon_\rho\right),
\quad
\rho_r^{(1)}=0,
\label{eq:epoch_update}
\end{equation}
where $\epsilon_\rho$ prevents $\rho_r=1$. If $\mathcal{V}_r^{(e)}$ is empty, the previous parameter is retained. Statistics are detached from backpropagation, and parameters remain fixed within each epoch.

Following t-vMF Dice Loss\cite{adaptivetvmf}, we minimize the squared deviation from one, the transformed similarity at perfect agreement:
\begin{equation}
\mathcal{L}_{\mathrm{STB}}
=
\frac{1}{BC}
\sum_{n=1}^{B}
\sum_{k=1}^{C}
\sum_{r\in\{b,i\}}
\left[
1-\phi\!\left(x_{n,k,r};\rho_r^{(e)}\right)
\right]^2,
\label{eq:stb_loss}
\end{equation}
where $B$ and $C$ are the batch size and number of classes. Empty regional targets give $x_{n,k,r}=0$ and thus a prediction-independent loss of $(1+\rho_r^{(e)})^2$; they are excluded only from the regional mean. The denominator in Eq.~\eqref{eq:tvmf_rho} is lower-bounded by $10^{-7}$ for numerical stability.

\begin{figure}[t]
\begin{center}
\includegraphics[scale=0.3]{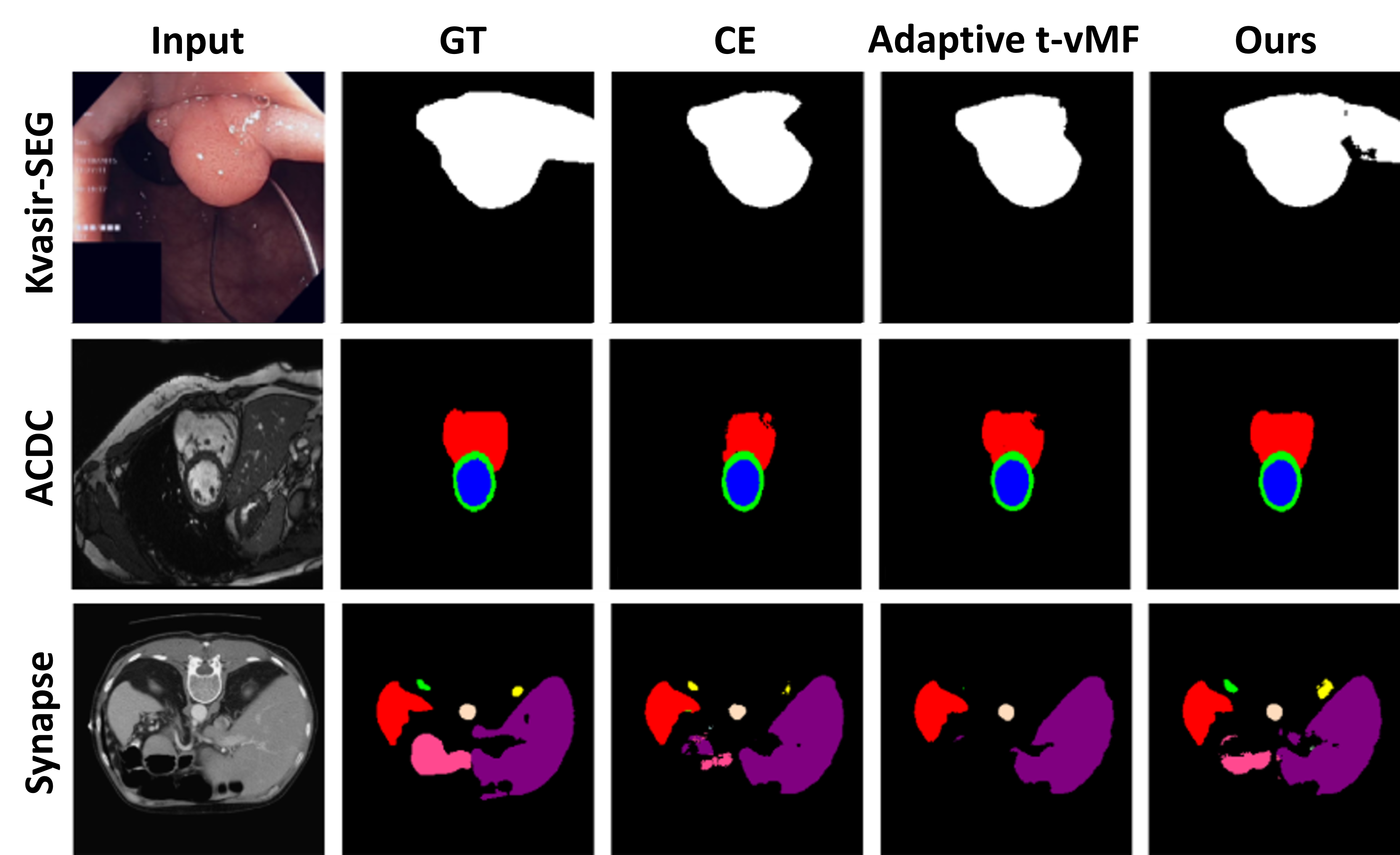}
\end{center}
\caption{
Qualitative results on three datasets using U-Net. From left to right, show Input image, Ground truth, Cross-entropy Loss, Adaptive t-vMF Dice Loss\cite{adaptivetvmf}, and STB-tvMF Dice Loss (Ours). For Adaptive t-vMF Dice Loss, we show predictions using the $\lambda$ with the highest DSC for each dataset.
}
\label{fig:vis}
\end{figure}

\section{Experiments}
\label{sec:experiments}

\subsection{Datasets and Metrics}

We evaluate the proposed method on three medical image segmentation datasets: Kvasir-SEG for polyp segmentation in endoscopic images\cite{kvasir}, ACDC for cardiac MRI segmentation\cite{acdc}, and Synapse for abdominal multi-organ CT segmentation\cite{smo}. 
The datasets contain 2, 4, and 9 classes, including background, respectively.

We conduct cross-validation using identical fold assignments across the compared losses. 
Segmentation performance is measured by the Dice similarity coefficient (DSC). 
We report the mean DSC across all classes, including background, averaged over all folds. 
Evaluation is performed on 2D images or slices.

\subsection{Training Conditions}

We evaluate all losses using U-Net\cite{unet}, TransUNet\cite{transunet}, and EMCAD\cite{emcad}. 
TransUNet uses the R50-ViT-B/16 configuration, while EMCAD uses a PVTv2-B2 encoder. 

We follow the training protocol of \cite{adaptivetvmf}, with architecture-specific batch sizes due to computational cost. 
All models are trained for 4,000 iterations using SGD
with a momentum of 0.9 and weight decay of $2\times10^{-4}$. 
The initial learning rate is 0.01 and follows a polynomial decay: $\eta_t=0.01(1-t/4000)^{0.9}$, where $t$ is the training iteration. 
The batch size is 24 for U-Net and 12 for TransUNet and EMCAD. 
Training images are resized to $224\times224$, augmented with random horizontal flipping and random rotation within $[-90^\circ,90^\circ]$, and normalized. 
At inference, images are resized to $224\times224$ and normalized.
Experiments are implemented in PyTorch and conducted on an NVIDIA RTX A6000 GPU.

We compare STB-tvMF Dice with Cross-entropy, Dice, Tversky, Focal Tversky, additive (CE + Dice) and multiplicative (CE $\times$ Dice) combinations of cross-entropy and Dice losses, TopK Dice, ABeDice, and Adaptive t-vMF Dice Loss\cite{dice, tversky, focaltversky, topkdice, abe, adaptivetvmf}. 
Baseline loss hyperparameters follow their original papers, and Adaptive t-vMF Dice Loss is evaluated with $\lambda \in \{2,32,128,256,1024\}$.
For STB-tvMF Dice, the boundary neighborhood is fixed at $3\times3$, and both shape parameters are initialized to zero and updated after each training epoch. 
No validation data are used to adapt the proposed loss.

\begin{table}[t]
\centering
\caption{Effect of adaptation granularity with epoch-wise updates. DSC (\%) using U-Net is shown. Global and class-wise variants use whole-image losses, whereas our method separates boundary and interior regions.}
\label{tab:ablation_granularity}
\scalebox{0.9}{
\begin{tabular}{lccc}
\bhline{1.5pt}
Granularity & Kvasir & ACDC & Synapse \\
\hline
Global ($1$)
& 91.21 & 91.54 & 52.41 \\
Class-wise ($C$)
& 91.54 & 91.80 & 47.38 \\
Boundary-interior ($2$)
& \textbf{91.91} & \textbf{91.92} & \textbf{70.75} \\
\bhline{1.5pt}
\end{tabular}
}
\end{table}

\begin{table}[t]
\centering
\caption{Effect of the shape-parameter update strategy in the boundary-interior formulation. DSC (\%) using U-Net is shown. The coefficient $m$ denotes EMA momentum.}
\label{tab:ablation_update}
\scalebox{0.9}{
\begin{tabular}{lccc}
\bhline{1.5pt}
Update strategy & Kvasir & ACDC & Synapse \\
\hline
Batch-wise
& 91.47 & \textbf{92.29} & 70.67 \\
EMA ($m=0.9$)
& 91.87 & 91.53 & 69.91 \\
EMA ($m=0.99$)
& 91.63 & 92.06 & 70.66 \\
Epoch-wise (Ours)
& \textbf{91.91} & 91.92 & \textbf{70.75} \\
\bhline{1.5pt}
\end{tabular}
}
\end{table}

\begin{table}[t]
\centering
\caption{Effect of the boundary extraction kernel size. Values are DSC (\%) using U-Net.}
\label{tab:ablation_kernel}
\scalebox{0.9}{
\begin{tabular}{lccc}
\bhline{1.5pt}
Kernel size & Kvasir & ACDC & Synapse \\
\hline
$3\times3$ (default)
& 91.91 & \textbf{91.92} & \textbf{70.75} \\
$5\times5$
& \textbf{92.09} & 91.47 & 70.29 \\
$7\times7$
& 91.81 & 91.45 & 68.69 \\
\bhline{1.5pt}
\end{tabular}
}
\end{table}

\subsection{Quantitative comparison}

Table~\ref{tab:main_results} summarizes segmentation performance across three datasets and three architectures. 
STB-tvMF Dice achieved the highest mean DSC in six of the nine settings and remained close to the best-performing baseline in the remaining three. 
On Synapse, it improved DSC over Dice loss by 5.73 and 3.56\% with U-Net and TransUNet, respectively. 
Compared with the best tested $\lambda$ of Adaptive t-vMF Dice Loss, the corresponding gains are 7.72 and 2.56\%. 
With EMCAD, it achieved the highest DSC on Kvasir-SEG and ACDC. 
These results demonstrate competitive performance across architectures using training statistics alone for shape adaptation, without validation-based shape updates or manual search for adaptation parameters.

\subsection{Qualitative comparison}
Fig.~\ref{fig:vis} presents selected segmentation examples. 
In the Kvasir-SEG example, STB-tvMF Dice recovered the rightward extension of the polyp that is largely missed by the two baselines. 
In the ACDC example, its predicted contours more closely followed the ground truth. 
In the Synapse example, it recovered structures that are missing or fragmented in the baseline predictions, although segmentation errors remained.

\subsection{Ablation studies}

Table~\ref{tab:ablation_granularity} uses the same self-tuning rule and epoch-wise updates, but changes how $\rho$ is assigned: one shared across all classes and pixels, one per class, or one each for boundary and interior regions. 
The boundary-interior formulation achieved the highest DSC on all three datasets. 
On Synapse, it outperformed the global and class-wise variants by 18.34 and 23.37\%, respectively. 
Pooling statistics across classes within each region may provide more reliable estimates than class-wise adaptation for infrequently occurring classes. 

Table~\ref{tab:ablation_update} fixed the boundary-interior formulation and compared four parameter-update strategies using U-Net: batch-wise updates, EMA with momentum coefficients of 0.9 and 0.99, and epoch-wise averaging. 
The DSC values differed by at most 0.84\% on each dataset. 
Epoch-wise averaging achieved the highest DSC on Kvasir-SEG and Synapse, whereas batch-wise updates performed best on ACDC. 
We adopted epoch-wise averaging because it maintained competitive accuracy across datasets without introducing an EMA momentum coefficient.

Table~\ref{tab:ablation_kernel} examines the boundary extraction kernel size using U-Net. 
DSC varies by less than 0.5 \% on Kvasir-SEG and ACDC, while increasing the kernel size from $3\times3$ to $7\times7$ reduces DSC by 2.06 \% on Synapse. 
These results support using a fixed $3\times3$ neighborhood across datasets.

\section{Conclusion}
\label{sec:conclusion}

We proposed STB-tvMF Dice, a loss function that separately adapts boundary and interior similarity functions using training statistics alone. 
Maximizing the local sensitivity of the function yields a simple update rule that sets each shape parameter to the preceding epoch's mean regional cosine similarity, eliminating manual search for adaptation parameters. 
Experiments on three datasets and three architectures demonstrated competitive segmentation performance, while ablation studies supported the boundary-interior formulation and epoch-wise updates. 
Future work will investigate its applicability to 3D segmentation and broader imaging settings.

\bibliographystyle{IEEEbib}
\bibliography{strings,refs}

\end{document}